\documentclass{article} 
\usepackage{nips13submit_e,times}
\usepackage{hyperref}
\usepackage{url}
\usepackage{graphicx}
\usepackage{amsmath,amssymb}
\usepackage{color}
\usepackage{tikz}
\usetikzlibrary{shapes,arrows,positioning}

\title{EL-GAN: Embedding Loss Driven Generative Adversarial Networks for Lane Detection}
\author{
Mohsen Ghafoorian, Cedric Nugteren, N\'ora Baka, Olaf Booij, Michael Hofmann \\
TomTom, Amsterdam, The Netherlands \\
{firstname.lastname}@tomtom.com \\
}

\newcommand{\elgan}{\mbox{EL-GAN}}
\newcommand\norm[1]{\left\lVert#1\right\rVert}
\nipsfinalcopy 

\begin{document}

\maketitle

\begin{abstract}

Convolutional neural networks have been successfully applied to semantic segmentation problems.
However, there are many problems that are inherently not pixel-wise classification problems but are nevertheless frequently formulated as semantic segmentation.
This ill-posed formulation consequently necessitates hand-crafted scenario-specific and computationally expensive post-processing methods to convert the per pixel probability maps to final desired outputs.
Generative adversarial networks (GANs) can be used to make the semantic segmentation network output to be more \emph{realistic} or better \emph{structure-preserving}, decreasing the dependency on potentially complex post-processing.

In this work, we propose \elgan{}: a GAN framework to mitigate the discussed problem using an \emph{embedding loss}.
With \elgan{}, we discriminate based on learned embeddings of both the labels and the prediction at the same time.
This results in much more stable training due to having better discriminative information, benefiting from seeing both `fake' and `real' predictions at the same time. This substantially stabilizes the adversarial training process.
We use the TuSimple lane marking challenge to demonstrate that with our proposed framework it is viable to overcome the inherent anomalies of posing it as a semantic segmentation problem.
Not only is the output considerably more similar to the labels when compared to conventional methods, the subsequent post-processing is also simpler and crosses the competitive 96\% accuracy threshold.

\end{abstract}

\section{Introduction}

Convolutional neural networks (CNNs) have been successfully applied to various computer vision problems by posing them as an image segmentation problem.
Examples include road scene understanding for autonomous driving~\cite{luc2016semantic,neven2018towards,pan2017spatial} and medical imaging~\cite{bentaieb2016topology,dai2017scan,huo2017splenomegaly,kohl2017adversarial,moeskops2017adversarial,oktay2017anatomically}.
The output of such a network is an image-sized map, representing per-pixel class probabilities.
However, in many cases the problem itself is not directly a pixel-classification task, and/or the predictions need to preserve certain qualities/structures that are not enforced with the high degrees of freedom of a per-pixel classification scheme. 
For instance, if the task at hand is to detect a single straight line in an image, a pixel-level loss cannot easily enforce high-level qualities such as thinness, straightness or the uniqueness of the detected line.
The fundamental reason behind this is the way the training loss is formulated (e.g. per-pixel cross entropy), such that each output pixel in the segmentation map is evaluated independently of all others, i.e. no explicit inter-pixel consistency is enforced.
Enforcing these qualities often necessitates additional post-processing steps.
Examples of post-processing steps include applying a conditional random field (CRF)~\cite{krahenbuhl2011efficient}, additional separately trained networks~\cite{neven2018towards}, or non-learned problem-specific algorithms~\cite{barhillel2014recent}.
Drawbacks of such approaches are that they require effort to construct, can have many hyper-parameters, are problem specific, and might still not capture the final objective.
For example, CRFs need to be trained separately and either only capture local consistencies or are computationally expensive at inference time with long-range dependencies.

A potential solution for the lack of structure enforcement in semantic segmentation problems is to use generative adversarial networks (GANs)~\cite{goodfellow2014generative} to `learn' an extra loss function that aims to model the desired properties.
GANs work by training two networks in an alternating fashion in a minimax game: a \textit{generator} is trained to produce results, while a \textit{discriminator} is trained to distinguish produced data (`fake') from ground truth labels (`real').
GANs have also been applied to semantic segmentation problems to try to address the aforementioned issues with the per-pixel loss~\cite{luc2016semantic}.
In such a case, the generator would produce the semantic segmentation map, while the discriminator alternately observes ground truth labels and predicted segmentation maps.
There are issues with this approach, as also observed by~\cite{xue2017segan}: the single binary prediction of the discriminator does not provide stable and sufficient gradient feedback to properly train the networks.

In prior work, the discriminator in a GAN observes either `real' or `fake' data in an alternating fashion (e.g.~\cite{luc2016semantic}), due to its inherently unsupervised nature.
However, in the case of a semantic segmentation problem, we do have access to the ground truth data corresponding to a prediction.
The intuition behind our work is that by feeding both the predictions and the labels at the same time, it is possible for a discriminator to obtain much more useful feedback to steer the training of the segmentation network in the direction of more realistic labels.
In other words, the discriminator can be taught to learn a supervised loss function.

In this work, we propose such an architecture for enforcing structure in semantic segmentation output.
In particular, we propose \elgan{} (`Embedding loss GAN'), in which the discriminator takes as input the source data, a prediction map and a ground truth label, and is trained to minimize the difference between embeddings of the predictions and labels.
The more useful gradient feedback and increased training stability in \elgan{} enables us to successfully train semantic segmentation networks using a GAN architecture.
As a result, our segmentation predictions are structurally much more similar to the training labels without requiring additional problem-specific loss terms or post-processing steps.
The benefits of our approach are illustrated in Fig.~\ref{fig:benefits_illustration}, in which we show an example training label and compare it to predictions of a regular segmentation network and our \elgan{} framework.
Our contributions are:

\begin{figure}
  \centering
  \includegraphics[width=0.31\columnwidth]{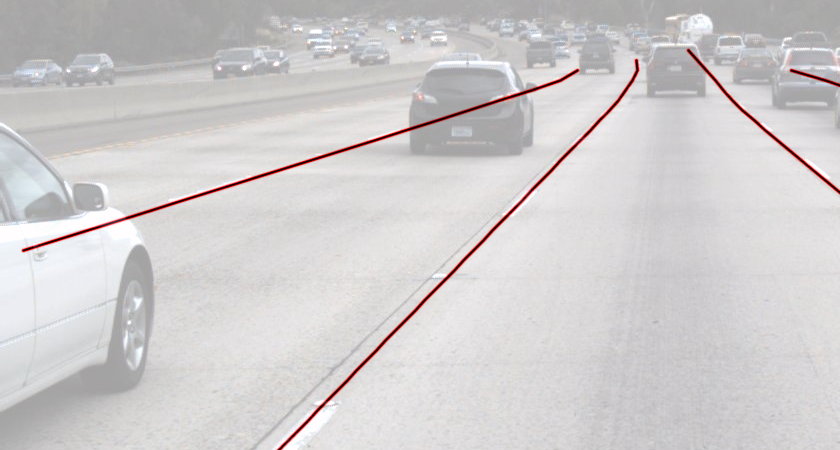}
  \includegraphics[width=0.31\columnwidth]{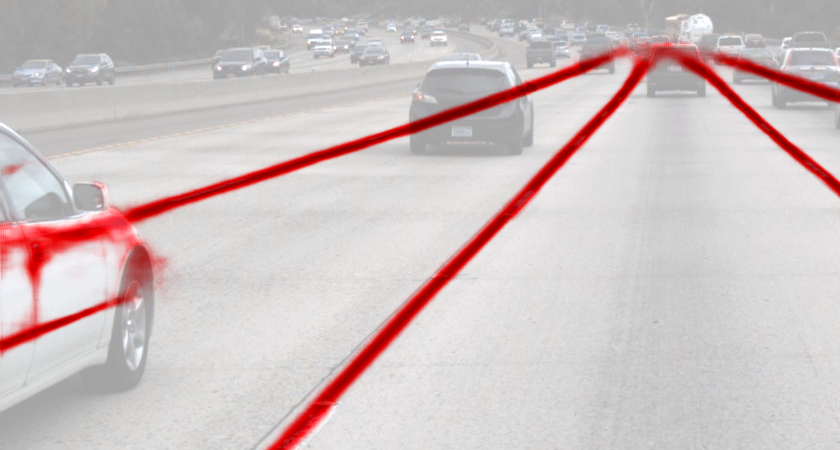}
  \includegraphics[width=0.31\columnwidth]{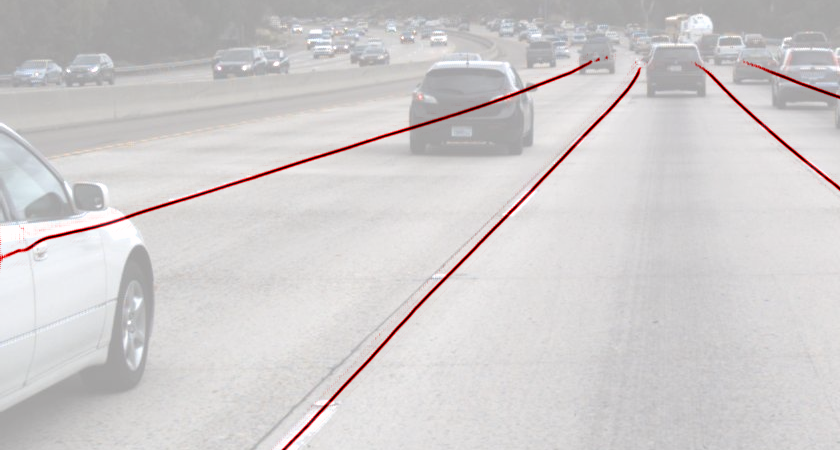}\\
  {\hspace{5em}label \hspace{8em} regular CNN \hspace{8em} \elgan{}\hspace{4em}}
  \caption{Illustration of using \elgan{} for lane marking segmentation: an example ground truth label (left), its corresponding raw prediction by a conventional segmentation network based on~\cite{jegou2017the} (middle), and a prediction by \elgan{} (right). Note how \elgan{} matches the thin-line style of the labels in terms of certainty and connectivity}
  \label{fig:benefits_illustration}
\end{figure}

\begin{itemize}

  \item
  We propose a novel method to impose structure on problems that are badly posed as semantic segmentation, by using a generative adversarial network architecture with a discriminator that is trained on both predictions and labels at the same time.
  We introduce \elgan{}, an instance of the above, which uses an $L_2$ loss on embeddings of the segmentation network predictions and the ground truth labels.

  \item
  We show that the embedding loss substantially stabilizes training and leads to more useful gradient feedback compared to a normal adversarial loss formulation.
  Compared to conventional segmentation networks, this requires no extra engineered loss terms or complex post-processing, leading to better label-like prediction qualities.

  \item
  We demonstrate the usefulness of \elgan{} for autonomous driving applications, although the method is generic and can be applied to other segmentation problems as well.
  We test on the TuSimple lane marking detection dataset and show competitive accuracy scores, but also show that \elgan{} visually produces results more similar to the ground truth labels.

\end{itemize}


\section{Related Work}


\textbf{Quality Preserving Semantic Segmentation.}
Several methods have proposed to add property-targeted loss terms \cite{bentaieb2016topology,oktay2017anatomically} or to use pair-wise or higher-order term CRFs \cite{krahenbuhl2011efficient,zheng2015conditional,schwing2015fully}, to enforce neural networks to preserve certain qualities such as smoothness, topology and neighborhood consistency. 
In contrast to our work, such approaches are mostly only capable of preserving lower-level consistencies and also impose additional costs at inference time.
Hand-engineering extra loss terms that target enforcing certain qualities is often challenging as identifying the target qualities in the first place and then coming up with efficient differentiable loss terms is often not straight-forward.

\textbf{Adversarial Training for Semantic Segmentation.}
The principal underlying idea of GANs~\cite{goodfellow2014generative} is to enable a neural network to learn a target distribution for generating samples by training it in a minimax game with a competing discriminator network.
Luc et al.~\cite{luc2016semantic} employed adversarial training for segmentation to ensure higher-level semantic consistencies.
Their work involves using a discriminator that provides feedback to the segmentation network (generator) based on differences between distributions of labels and predictions.
This differs from aforementioned works in the sense that the additional loss term is being learned by the discriminator rather than having fixed hand-crafted loss terms.
The same mechanism was later applied to image-to-image translation~\cite{isola2017image}, medical image analysis~\cite{dai2017scan,huo2017splenomegaly,kohl2017adversarial,moeskops2017adversarial,xue2017segan,yang2017automatic,li2017brain,sadanandan4spheroid} and other segmentation tasks~\cite{nguyen2017shadow}.
In contrast to our work, this formulation of adversarial training does not use the pairing information of images and labels.
Based on this, some works~\cite{zhang2017deep,hung2018adversarial} suggest using a GAN in a semi-supervised fashion, with the additional assumption that the unlabeled data is coming from the same distribution as the labeled ones.
Our work also stems from the same intuition that this formulation does not leverage the pairing information; we instead change the method such that the pairing information is exploited.
Another related work is~\cite{xue2017segan}, which proposes an $L_1$ loss term for GAN-based medical image segmentation, but interpretations and extensive ablation studies are not provided.
Our method differs in the input the discriminator receives, as well as the loss term that is used to train it.
In concurrent work, Hwang et al.~\cite{hwang2018adversarial} uses adversarial training for structural matching between the ground-truth and the predicted image.
In contrast to our work,~\cite{hwang2018adversarial} does not condition the discriminator on the input image, nor uses a pixel-level loss to steer the training of the segmenter network. 
As a consequence, the discriminator representations need to be kept low-level to ensure a segmenter that attends to low-level details. 
Furthermore, we provide extensive ablation studies in order to better understand, discuss and interpret the characteristics and benefits of the method.

\textbf{Perceptual Loss.}
Several recent works~\cite{sajjadi2017enhancenet,dosovitskiy2016generating,johnson2016perceptual}, in particular targeting image super-resolution, are based on the idea that pixel-level objective losses are often not sufficient to ensure high-level semantics of a generated image.
Therefore, they suggest to capture higher-level representations of images from the representations of a separate network at a given layer.
In image super-resolution, the corresponding ground truth label for a given low-resolution image is often available.
Therefore, a difference measure (e.g. $L_2$) between the high-level representations of the reconstructed and ground truth images is considered as an extra loss term.
Our work is inspired by this idea: similarly, we propose to use the difference between the labels and predictions in a high-level embedding space.

\textbf{Lane Marking Detection.}
Since the evaluation of our work focuses on lane marking detection, we also discuss other related approaches for this problem, while we refer the reader to a recent survey for a broader overview~\cite{barhillel2014recent}.
An example of a successful lane marking detection approach is by Pan et al.~\cite{pan2017spatial}.
In their work, they train a problem-specific spatial CNN and add hand-crafted post-processing.
Lee et al.~\cite{lee2017vpgnet} use extra vanishing-point labels to guide the network toward a more structurally consistent lane marking detection.
Another recent example is the work by Neven et al.~\cite{neven2018towards}, in which a regular segmentation network is used to obtain lane marking prediction maps.
They then train a second network to perform a constrained perspective transformation, after which curve fitting is used to obtain the final results.
We compare our work in more detail to the studies above \cite{neven2018towards,pan2017spatial} that are similarly conducted on the Tusimple challenge, in Section~\ref{sec:tusimple_comparison}.


\section{Method}
\label{sec:method}

In this section we introduce \elgan{}: adversarial training with embedding loss for semantic segmentation.
This method is generic and can be applied to various segmentation problems.
The detailed network architecture and training set-up is discussed in Section~\ref{sec:experimental_setup}.


\subsection{Baseline: Adversarial Training for Semantic Segmentation}

Adversarial training can be used to ensure a higher level of label resembling qualities such as smoothness, preserving neighborhood consistencies, and so on.
This is done by using a discriminator network that learns a loss function for these desirable properties over time rather than formulating these properties explicitly.
A typical approach for benefiting from adversarial training for semantic segmentation \cite{luc2016semantic,isola2017image} involves formulating a loss function for the segmentation network (generator) that consists of two terms:
one term concerning low-level pixel-wise prediction/label fitness ($\mathcal{L}_\text{fit}$) and another (adversarial) loss term for preserving higher-level consistency qualities ($\mathcal{L}_\text{adv}$), conditioned on the input image:
\begin{align}
\mathcal{L}_\text{gen}(x, y; \theta_\text{gen}, \theta_\text{disc}) = \mathcal{L}_\text{fit}(G(x; \theta_\text{gen}), y) + \lambda \mathcal{L}_\text{adv}(G(x; \theta_\text{gen}); x, \theta_\text{disc}),
\label{eq:lgen_regular}
\end{align}
where $x$ and $y$ are the input image and the corresponding label map respectively, $\theta_\text{gen}$ and $\theta_\text{disc}$ are the set of parameters for the generator and discriminator networks, $G(x;\theta)$ represents a transformation on input image $x$, imposed by the generator network parameterized by $\theta$, and $\lambda$ indicates the relative importance of the adversarial loss term.
The loss term $\mathcal{L}_\text{fit}$ is often formulated with a pixel-wise categorical cross entropy loss, $\mathcal{L}_\text{cce}(G(x; \theta_\text{gen}), y)$, where $\mathcal{L}_\text{cce}(\hat{y}, y) = \frac{1}{wh}\sum_{i}^{wh} \sum_{j}^{c} y_{i,j} \ln(\hat{y}_{i, j})$ with $c$ representing the number of target classes and $w$ and $h$ being the width and height of the image.

The adversarial loss term, $\mathcal{L}_\text{adv}$ indicates how successful the discriminator is in rejecting the (fake) dense prediction maps produced by the generator and is often formulated with a binary cross entropy loss between zero and the binary prediction of the discriminator for a generated prediction map: $\mathcal{L}_\text{bce}(D(G(x; \theta_\text{gen});\theta_\text{disc}), 0)$, where $\mathcal{L}_\text{bce}(\hat{z}, z) = -z \ln(\hat{z})-(1-z) \ln(1-\hat{z})$ and $D$ is the transformation imposed by the discriminator network.

While the generator is trained to minimize its adversarial loss term, the discriminator tries to maximize it, by minimizing its loss defined as:
\begin{align}
\mathcal{L}_\text{disc}(x, y; \theta_\text{gen}, \theta_\text{disc}) = \mathcal{L}_\text{bce}(D(G(x; \theta_\text{gen}); \theta_\text{disc}), 1) + \mathcal{L}_\text{bce}(D(y; \theta_\text{disc}), 0).
\label{eq:ldisc_ce}
\end{align}
By alternating between the training of the two networks, the discriminator learns the differences between the label and prediction distributions, while the generator tries to change the qualities of its predictions, similar to that of the labels, such that the two distributions are not distinguishable.
In practice, it is often observed that the training of the adversarial networks tends to be more tricky and unstable compared to training normal networks.
This can be attributed to the mutual training of the two networks involved in a minimax game where the training dynamics of each affect the training of the other.
The discriminator gives feedback to the generator based on how plausible the generator images are.
There are two important issues with the frequently used adversarial training framework for semantic segmentation:
\begin{enumerate}
  \item
  The notion of plausibility and fake-ness of these prediction maps comes from the discriminator's representation of these concepts and how its weights encode these qualities;
  This encoding is likely to be far from perfect, resulting in gradients in directions that are likely not improving the generator.
  \item
  The conventional adversarial loss term does not exploit the valuable piece of information on image/label pairing that is often available for many of the supervised semantic segmentation tasks.
\end{enumerate}


\subsection{Adversarial Training with Embedding Loss}

Given the two issues above, one can leverage the image/label pairing to base the plausibility/fake-ness not only on the discriminator's understanding of these notions but also on a true plausible label map.
One way to utilize this idea is to use the discriminator to take the prediction/label maps into a higher-level description and define the adversarial loss as their difference in embedding space:
\begin{align}
\mathcal{L}_\text{gen}(x, y; \theta_\text{gen}, \theta_\text{disc}) = \mathcal{L}_\text{fit}(G(x; \theta_\text{gen}), y) + \lambda \mathcal{L}_\text{adv}(G(x; \theta_\text{gen}), y; x, \theta_\text{disc}),
\label{eq:lgen_twoinputs}
\end{align}
where we suggest to formulate $\mathcal{L}_\text{adv}(G(x; \theta_\text{gen}), y; x, \theta_\text{disc})$ with embedding loss $\mathcal{L}_\text{emb}(G(x; \theta_\text{gen}), y; x, \theta_\text{disc})$ defined as a distance over embeddings (e.g. $L_2$):
\begin{align}
\mathcal{L}_\text{emb}(\hat{y}, y; x, \theta_\text{disc})=\norm{D_{e}(y;x,\theta_\text{disc})-D_{e}(\hat{y}; x, \theta_\text{disc})}_2,
\label{eq:lgen_embeddingloss}
\end{align}
where $D_{e}(\hat{y}; x, \theta)$ represents the embeddings extracted from a given layer in the network $D$ parameterized with $\theta$, given the prediction $\hat{y}$ and $x$ as its inputs.

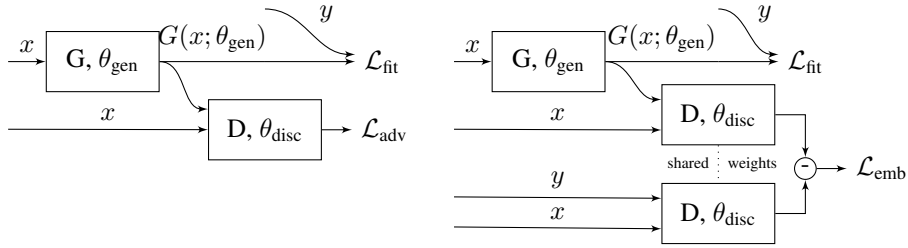
\begin{figure}[t]
  \centering
  \tikzstyle{network} = [draw, rectangle, minimum height=0.8cm, minimum width=1.5cm]
  \tikzstyle{concat} = [draw, circle, inner sep=0.05cm]
  \tikzstyle{loss} = [node distance=0.5cm]
  \tikzstyle{point} = [coordinate]
  \begin{tikzpicture}[auto, node distance=2cm,>=latex']
      \node [point, name=xin1a] {};
      \node [point, name=xin1b, below=0.9cm of xin1a] {};
      \node [network, name=generator1, right=0.5cm of xin1a] {G, $\theta_\text{gen}$};
      \node [point, name=prediction1, right=1.4cm of generator1] {};
      \node [network, name=discriminator1, below=0.5cm of prediction1] (discriminator1) {D, $\theta_\text{disc}$};
      \node [point, name=yin1, above=0.7cm of prediction1] {};
      \node [loss, name=lossadv1, right=0.4cm of discriminator1] {$\mathcal{L}_\text{adv}$};
      \node [loss, name=lossfit1, right=1.2cm of prediction1] {$\mathcal{L}_\text{fit}$};
      \draw [draw,->] (xin1a) -- node {$x$} (generator1);
      \draw [draw,->] (xin1b) -- node {$x$} (discriminator1);
      \draw [] (generator1) edge[out=0, in=180,->] node {} (discriminator1.160);
      \draw [-] (generator1) -- node {$G(x; \theta_\text{gen})$} (prediction1);
      \draw [->] (prediction1) -> node {} (lossfit1);
      \draw [] (yin1) edge[out=0, in=180,->] node {$y$} (lossfit1.165);
      \draw [->] (discriminator1) -- node {} (lossadv1);
      \node [point, name=separator, right=0.3cm of lossfit1] {};
      \node [point, name=xin2a, right=0.3cm of separator] {};
      \node [point, name=xin2b, below=0.9cm of xin2a] {};
      \node [point, name=yin2b, below=0.9cm of xin2b] {};
      \node [point, name=xin2c, below=0.4cm of yin2b] {};
      \node [network, name=generator2, right=0.5cm of xin2a] {G, $\theta_\text{gen}$};
      \node [point, name=prediction2, right=1.5cm of generator2] {};
      \node [network, name=discriminator2a, below=0.3cm of prediction2] (discriminator2a) {D, $\theta_\text{disc}$};
      \node [network, name=discriminator2b, below=0.5cm of discriminator2a] (discriminator2b) {D, $\theta_\text{disc}$};
      \node [point, name=yin2a, above=0.7cm of prediction2] {};
      \node [loss, name=lossfit2, right=0.8cm of prediction2] {$\mathcal{L}_\text{fit}$};
      \node [concat, name=lossembedding, below=1.0cm of lossfit2] {-};
      \node [loss, name=lossadv2, right=0.4cm of lossembedding] {$\mathcal{L}_\text{emb}$};
      \draw [draw,->] (xin2a) -- node {$x$} (generator2);
      \draw [draw,->] (xin2b) -- node {$x$} (discriminator2a.195);
      \draw [draw,->] (yin2b) -- node {$y$} (discriminator2b.165);
      \draw [draw,->] (xin2c) -- node {$x$} (discriminator2b.195);
      \draw [draw,dotted] (discriminator2a) -- node [left, pos=0.45] {\tiny shared} node [right, pos=0.5] {\tiny weights} (discriminator2b);
      \draw [] (generator2) edge[out=0, in=180,->] node {} (discriminator2a.165);
      \draw [-] (generator2) -- node {$G(x; \theta_\text{gen})$} (prediction2);
      \draw [->] (prediction2) -> node {} (lossfit2);
      \draw [-] (yin2a) edge[out=0, in=180,->] node {$y$} (lossfit2.165);
      \draw [->] (discriminator2a) -| node {} (lossembedding);
      \draw [->] (discriminator2b) -| node {} (lossembedding);
      \draw [->] (lossembedding) -- node {} (lossadv2);
  \end{tikzpicture}
  \caption{Illustration of the novel training set-up for the generator loss: left for a conventional GAN (Equation~\ref{eq:lgen_regular}), right when using the embedding loss (Equations~\ref{eq:lgen_twoinputs} and~\ref{eq:lgen_embeddingloss})}
  \label{fig:gan_setup}
\end{figure}

We name this the \elgan{} architecture, in which the adversarial loss and the corresponding gradients are computed based on a difference in high-level descriptions (embeddings) of labels and predictions.
While the discriminator learns to minimize its loss on the discrimination between real and fake distributions, and likely learns a set of discriminative embeddings, the generator tries to minimize this embedding difference.
This formulation of generator training is illustrated in Fig.~\ref{fig:gan_setup} on the right-hand side, in which we also show the regular generator training set-up on the left-hand side for comparison.

Apart from the mentioned change in computing the adversarial loss for the generator updates, Equation~\ref{eq:ldisc_ce} for discriminator updates can optionally be rewritten with a similar idea as:
\begin{align}
\mathcal{L}_\text{disc}(x, y; \theta_\text{gen}, \theta_\text{disc}) = - \mathcal{L}_\text{emb}(G(x,\theta_\text{gen}),y;x,\theta_\text{disc}).
\label{eq:ldisc_emb}
\end{align}
However, in our empirical studies we have found that using the cross entropy loss for updating the discriminator parameters gives better results.


\section{Experimental Setup}
\label{sec:experimental_setup}

In this section we elaborate on the datasets and metrics used for evaluating our method, followed by details of the network architectures and training methods.


\subsection{Evaluation Datasets and Metrics}

We focus our evaluation on the application domain of autonomous driving, but stress that our method is generic and can be applied to other semantic segmentation problems as well.
One of the motivations of this work is to be able to produce predictions resembling the ground truth labels as much as possible.
This is in particular useful for the TuSimple lane marking detection data set with thin structures, reducing the need for complicated post-processing.


The TuSimple lane marking detection dataset\footnote{TuSimple dataset details: \url{http://benchmark.tusimple.ai/\#/t/1}} consists of 3626 annotated 1280$\times$720 front-facing road images images on US highways in the San Diego area divided over four sequences, and a similar set of 2782 test images.
The annotations are given in the form of polylines of lane markings: those of the ego-lane and the lanes to the left and right of the car.
The polylines are given at fixed height-intervals every 20 pixels.
To generate labels for semantic segmentation, we convert these to segmentation maps by discretizing the lines using smooth interpolation with a Gaussian with a sigma of 1 pixel wide.
An example of such a label is shown in red in the left of Fig.~\ref{fig:benefits_illustration}.

The dataset is evaluated on results in the same format as the labels, namely multiple polylines.
For our evaluation we use the official metrics as defined in the challenge$^1$, namely accuracy, false positive rate, and false negative rate.
We report results on the official test set as well as on a validation set which is one of the labeled sequences with 409 images (`0601').
We note that performance on this validation set is perhaps not fully representative, because of its small size.
A different validation sequence also has its drawbacks, since the other three are much larger and will considerably reduce the size of the already small data set.

Since our network still outputs segmentation maps rather than the required polylines, we do apply post-processing, but keep it as simple as possible: after binarizing, we transform each connected component into a separate polyline by taking the mean x-index of a sequence of non-zero values at each y-index.
We refer to this method as `basic'.
We also evaluate a `basic++' version which also splits connected components in case it detects that multiple sequences of non-zero values occur at one sampling location.





\subsection{Network Architectures and Training}

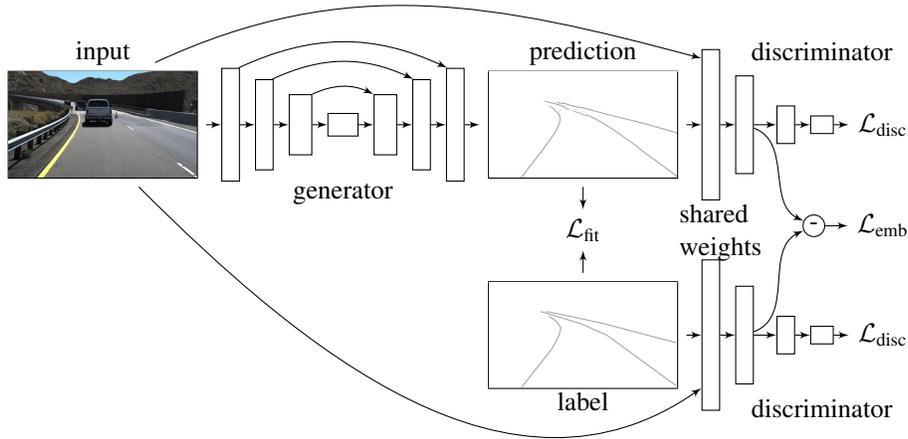
\begin{figure}[!t]
  \centering
  \tikzstyle{layer} = [draw, rectangle, node distance=0.2cm]
  \tikzstyle{concat} = [draw, circle, inner sep=0.05cm]
  \tikzstyle{loss} = [node distance=0.5cm]
  \tikzstyle{description} = []
  \tikzstyle{image} = []
  \tikzstyle{point} = [coordinate]
  \begin{tikzpicture}[auto, node distance=2cm,>=latex']
      \node [image, name=imagein, label={[label distance=-0.15cm]input}]
            {\frame{\includegraphics[width=2.5cm]{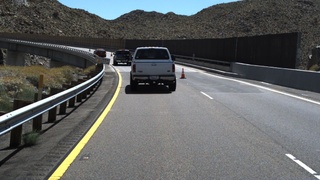}}};
      \node [layer, name=genl1, right=of imagein, minimum height=1.5cm, minimum width=0.05cm] {};
      \node [layer, name=genl2, right=of genl1, minimum height=1.2cm, minimum width=0.15cm] {};
      \node [layer, name=genl3, right=of genl2, minimum height=0.8cm, minimum width=0.3cm] {};
      \node [layer, name=genl4, right=of genl3, minimum height=0.3cm, minimum width=0.4cm] {};
      \node [layer, name=genl5, right=of genl4, minimum height=0.8cm, minimum width=0.3cm] {};
      \node [layer, name=genl6, right=of genl5, minimum height=1.2cm, minimum width=0.15cm] {};
      \node [layer, name=genl7, right=of genl6, minimum height=1.5cm, minimum width=0.05cm] {};
      \node [image, name=prediction, right=0.2cm of genl7, label={[label distance=-0.15cm]prediction}]
            {\frame{\includegraphics[width=2.5cm]{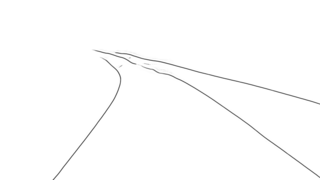}}};
      \node [loss, name=lossfit, below=0.3cm of prediction] {$\mathcal{L}_\text{fit}$};
      \node [image, name=labelmap, below=0.3cm of lossfit, label={[label distance=-1.95cm]label}]
            {\frame{\includegraphics[width=2.5cm]{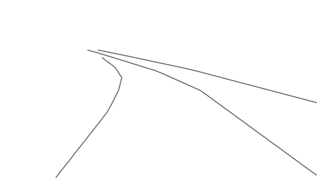}}};
      \node [layer, name=discl1a, right=of prediction, minimum height=2cm, minimum width=0.05cm] {};
      \node [layer, name=discl2a, right=of discl1a, minimum height=1.3cm, minimum width=0.1cm] {};
      \node [layer, name=discl3a, right=0.3cm of discl2a, minimum height=0.5cm, minimum width=0.2cm] {};
      \node [layer, name=discl4a, right=of discl3a, minimum height=0.2cm, minimum width=0.3cm] {};
      \node [loss, name=disclossa, right=0.2cm of discl4a] {$\mathcal{L}_\text{disc}$};
      \node [loss, name=embloss, below=0.8cm of disclossa] {$\mathcal{L}_\text{emb}$};
      \node [layer, name=discl1b, right=of labelmap, minimum height=2cm, minimum width=0.05cm] {};
      \node [layer, name=discl2b, right=of discl1b, minimum height=1.3cm, minimum width=0.1cm] {};
      \node [layer, name=discl3b, right=0.3cm of discl2b, minimum height=0.5cm, minimum width=0.2cm] {};
      \node [layer, name=discl4b, right=of discl3b, minimum height=0.2cm, minimum width=0.3cm] {};
      \node [loss, name=disclossb, right=0.2cm of discl4b] {$\mathcal{L}_\text{disc}$};
      \node [concat, name=minus, left=0.3cm of embloss] {-};
      \node [description, name=gentext, below=0.5cm of genl4] {generator};
      \node [description, name=disctext1, above=0.6cm of discl4a] {discriminator};
      \node [description, name=disctext2, below=0.6cm of discl4b] {discriminator};
      \node [description, name=sharedweights1, above right=-0.3cm and 0.8cm of lossfit] {shared};
      \node [description, name=sharedweights1, below right=-0.3cm and 0.8cm of lossfit] {weights};
      \draw [draw,->] (imagein) -- node {} (genl1);
      \draw [draw,->] (genl1) -- node {} (genl2);
      \draw [draw,->] (genl2) -- node {} (genl3);
      \draw [draw,->] (genl3) -- node {} (genl4);
      \draw [draw,->] (genl4) -- node {} (genl5);
      \draw [draw,->] (genl5) -- node {} (genl6);
      \draw [draw,->] (genl6) -- node {} (genl7);
      \draw [draw] (genl1.82) edge[out=25, in=155,->] node {} (genl7.98);
      \draw [draw] (genl2.80) edge[out=25, in=155,->] node {} (genl6.100);
      \draw [draw] (genl3.69) edge[out=25, in=155,->] node {} (genl5.111);
      \draw [draw,->] (genl7) -- node {} (prediction);
      \draw [draw,->] (prediction) -- node {} (lossfit);
      \draw [draw,->] (labelmap) -- node {} (lossfit);
      \draw [draw,->] (prediction) -- node {} (discl1a);
      \draw [draw,->] (discl1a) -- node {} (discl2a);
      \draw [draw,->] (discl2a) -- node {} (discl3a);
      \draw [draw,->] (discl3a) -- node {} (discl4a);
      \draw [draw,->] (discl4a) -- node {} (disclossa);
      \draw [draw] (discl2a) edge[out=340, in=155,->] node {} (minus);
      \draw [draw,->] (labelmap) -- node {} (discl1b);
      \draw [draw,->] (discl1b) -- node {} (discl2b);
      \draw [draw,->] (discl2b) -- node {} (discl3b);
      \draw [draw,->] (discl3b) -- node {} (discl4b);
      \draw [draw,->] (discl4b) -- node {} (disclossb);
      \draw [draw] (discl2b) edge[out=20, in=205,->] node {} (minus);
      \draw [draw,->] (minus) -- node {} (embloss);
      \draw [draw] (imagein.50) edge[out=25, in=165,->] node {} (discl1a.98);
      \draw [draw] (imagein.300) edge[out=315, in=215,->] node {} (discl1b.260);
  \end{tikzpicture}
  \vspace{-1cm}
  \caption{Overview of the \elgan{} architecture, illustrating both the training of the generator and discriminator with examples from the TuSimple lane marking challenge}
  \label{fig:elgan_tusimple}
\end{figure}

In this section we discuss the network and training set-up used for our experiments.
A sketch of the high-level network architecture with example data is shown in Fig.~\ref{fig:elgan_tusimple}, which shows the different loss terms used for either the generator or discriminator training, or both.

For the generator we use a fully-convolutional U-Net style network with a downwards and an upwards path and skip connections.
In particular, we use the Tiramisu DenseNet architecture~\cite{jegou2017the} for lane marking detection, configured with 7 up/down levels for a total of 64 3$\times$3 convolution layers.

For the discriminator we use a DenseNet architecture~\cite{huang2016densely} with 7 blocks and a total of 32 3$\times$3 convolution layers, followed by a fully-convolutional \textit{patch-GAN} classifier~\cite{li2016precomputed}.
We use a two-headed network for the first 2 dense blocks to separately process the input image from the labels or predictions, after which we concatenate the feature maps.
We take the embeddings after the final convolution layer, but explore other options in Section~\ref{sec:ablation_studies}.

We first pre-train the generator models until convergence, which we also use as our baseline non-GAN model for Section~\ref{sec:results}.
Using a batch size of 8, we then pre-train the discriminator for 10k iterations, after which alternate between 300 and 200 iterations of generator and discriminator training, respectively.
The generator is trained with the Adam optimizer, while the discriminator training was observed to be more stable using SGD.
We train the discriminator using the regular cross entropy loss (Equation~\ref{eq:ldisc_ce}), while we train the generator with the adversarial embedding loss with $\lambda=1$ (Equations~\ref{eq:lgen_twoinputs} and~\ref{eq:lgen_embeddingloss}).
We did not do any data augmentation nor pre-train the model on other data.


\section{Results}
\label{sec:results}

In this section we report the results on the TuSimple datasets using the experimental set-up as discussed in Section~\ref{sec:experimental_setup}.
Additionally, we perform three ablation studies: evaluating the training stability, exploring the options for the training losses, and varying the choice for embedding loss layer.


\subsection{TuSimple Lane Marking Challenge}

In this section we report the results of the TuSimple lane marking detection challenge and compare them with our baseline and the state-of-the-art.

We first evaluated \elgan{} and our baseline on the validation set using both post-processing methods.
The results in Table~\ref{table:tusimple_simple_pp} show that the basic post-processing method is not suitable for the baseline model, while the improved basic++ method performs a lot better.
Still, \elgan{} outperforms the baseline, in particular with the most basic post-processing method.

\begin{figure}[!t]
  {
  \centering
  \includegraphics[width=1.0\columnwidth]{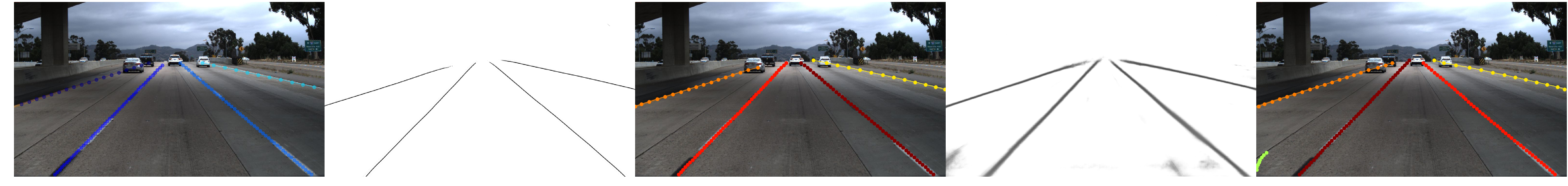}
  \includegraphics[width=1.0\columnwidth]{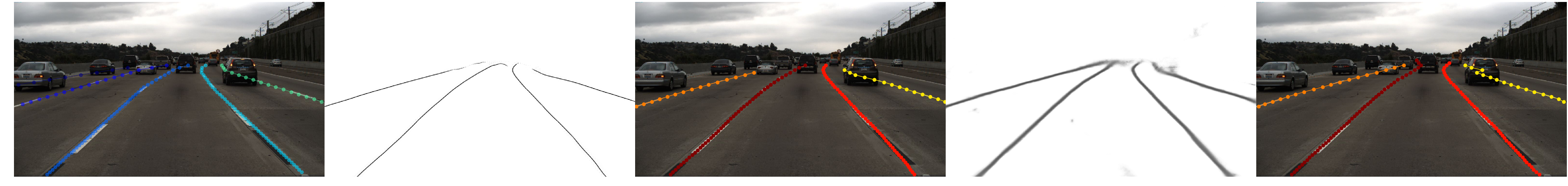}
  \includegraphics[width=1.0\columnwidth]{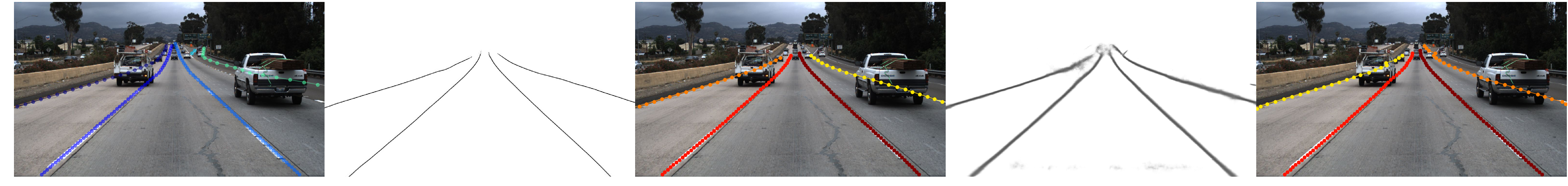}
  \includegraphics[width=1.0\columnwidth]{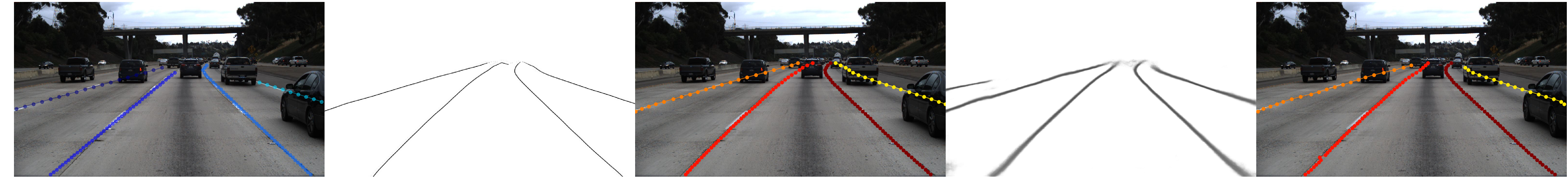}
  \includegraphics[width=1.0\columnwidth]{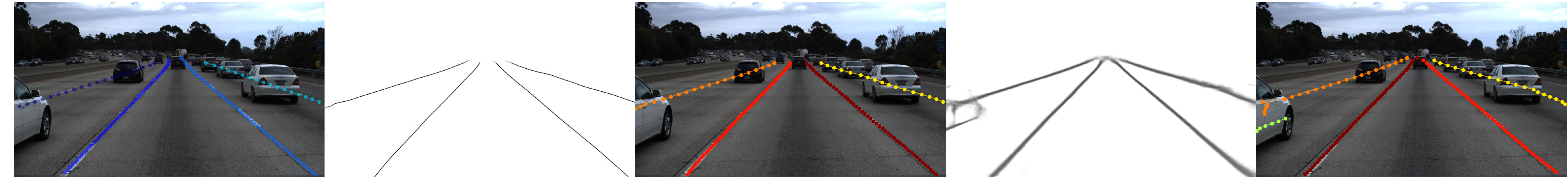}
  \includegraphics[width=1.0\columnwidth]{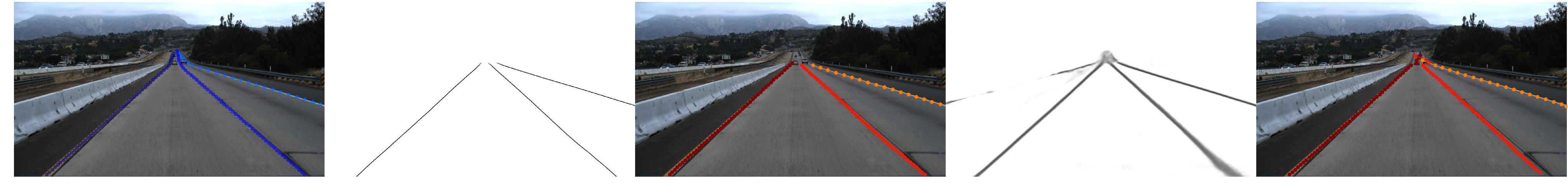}
  }
  {\phantom{} \hspace{1.5em} label on data \hspace{2.5em} prediction \hspace{2.5em} post-processed \hspace{2em} prediction \hspace{2.5em} post-processed \hspace{0em}}
  \\
  {\phantom{} \hspace{14em} \elgan{} \hspace{11em} regular CNN \hspace{1em}}
  \caption{Example results for lane marking segmentation: the labels on top of the data (left column), the prediction and final results of \elgan{} (next two columns), and results of the regular CNN baseline~\cite{jegou2017the} using the same post-processing (right two columns). The colors of the lines have no meaning other than to distinguish them from each other.
  Details are best viewed on a computer screen when zoomed in
  }
  \label{fig:tusimple_simple_pp}
\end{figure}

\setlength{\tabcolsep}{4pt}
\begin{table}
  \begin{center}
    \caption{Results on TuSimple lane marking validation set}
    \label{table:tusimple_simple_pp}
    \begin{tabular}{ll|ccc}
      \hline\noalign{\smallskip}
      Method & Post-processing & Accuracy (\%) & FP & FN \\
      \noalign{\smallskip}\hline\noalign{\smallskip}
      Baseline (no GAN) & basic & 86.2 & 0.089 & 0.213 \\
      Baseline (no GAN) & basic++ & 94.3 & 0.084 & 0.070 \\
      \elgan{} & basic & 93.3 & 0.061 & 0.104 \\
      \elgan{} & basic++ & \textbf{94.9} & \textbf{0.059} & \textbf{0.067} \\
      \noalign{\smallskip}\hline
    \end{tabular}
  \end{center}
\end{table}
\setlength{\tabcolsep}{1.4pt}

Some results on the validation set are shown in Fig.~\ref{fig:tusimple_simple_pp}, which compares the two methods in terms of raw prediction maps and post-processed results using the basic++ method.
Clearly, \elgan{} produces considerably thinner and more label-like output with less noise, making post-processing easier in general.

\setlength{\tabcolsep}{4pt}
\begin{table}[ht]
  \begin{center}
    \caption{TuSimple lane marking challenge leaderboard (test set) as of March 14, 2018}
    \label{table:tusimple_simple_leaderboard}
    \begin{tabular}{c|llc|ccc}
      \hline\noalign{\smallskip}
      Rank & Method & Name on board & Extra data & Accuracy (\%) & FP & FN\\
      \noalign{\smallskip}\hline\noalign{\smallskip}
      \#1 & Unpublished & leonardoli & ? &\textbf{96.87} & 0.0442 & 0.0197 \\
      \#2 & Pan et al.~\cite{pan2017spatial} & XingangPan & Yes & 96.53 & 0.0617 & \textbf{0.0180} \\
      \#3 & Unpublished & aslarry & ? & 96.50 & 0.0851 & 0.0269 \\
      \#5 & Neven et al.~\cite{neven2018towards} & DavyNeven & No & 96.38 & 0.0780 & 0.0244 \\
      \#6 & Unpublished & li & ? & 96.15 & 0.1888 & 0.0365 \\
      \noalign{\smallskip}\hline\noalign{\smallskip}
      \#14 & Baseline (no GAN) & N/A  & No & 94.54 & 0.0733 & 0.0476 \\
      \#4 & \elgan{} & TomTom \elgan{} & No & 96.39 & \textbf{0.0412} & 0.0336 \\
      \noalign{\smallskip}\hline
    \end{tabular}
  \end{center}
\end{table}
\setlength{\tabcolsep}{1.4pt}

Furthermore, we train \elgan{} and the baseline on the entire labeled dataset, and evaluate using the basic++ post-processing on the official test set of the TuSimple challenge.
Table~\ref{table:tusimple_simple_leaderboard} shows the results, which includes all methods in the top 6 (only two of which are published, to the best of our knowledge) and their rank on the leaderboard as of March 14, 2018.
We rank 4th based on accuracy with a difference less than half a percent to the best, and obtain the lowest false positive rate.
Compared to the baseline, our adversarial training algorithm improves $\sim$2\% on the accuracy (decrease of error by 38\%), decreases the FPs by more than 55\% and FNs by 30\% on the private challenge test set. 
These improvements take the baseline from 14th rank to 4th.

\subsection{Ablation Studies}
\label{sec:ablation_studies}

\begin{figure}[t]
  \centering
  \includegraphics[width=0.99\columnwidth]{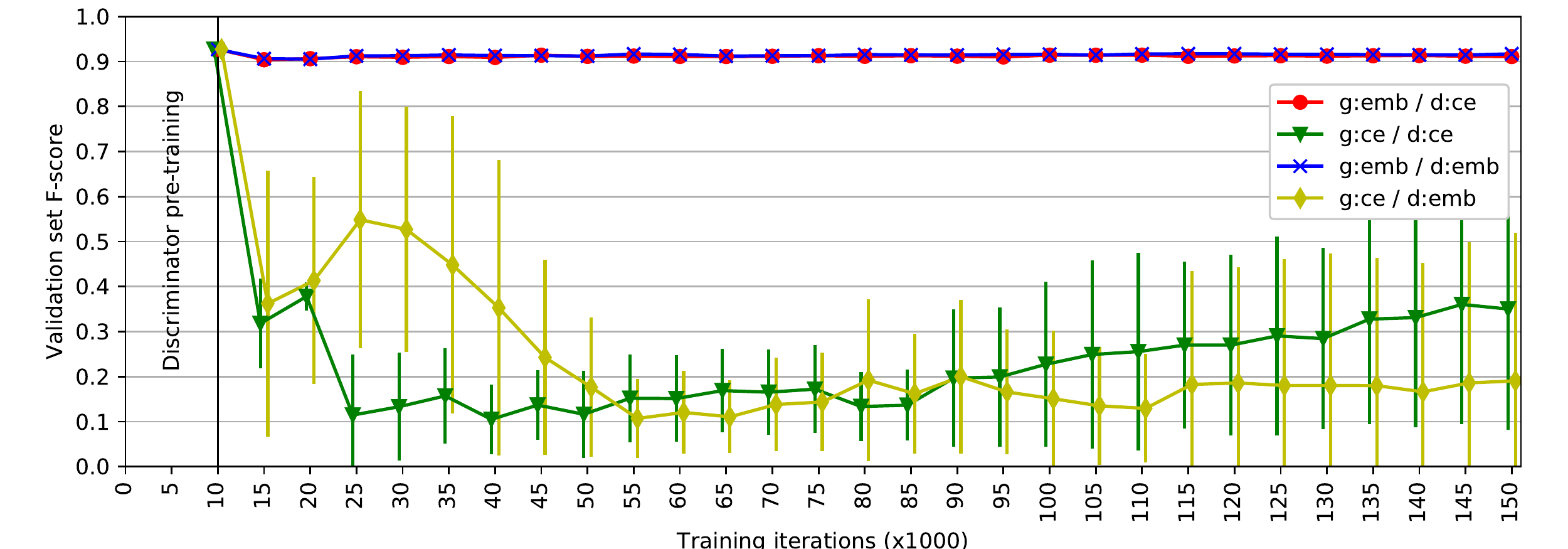}
  \caption{A comparison of training stability for using different adversarial loss terms (embedding/cross entropy) on the validation f-score. For each method the central point represents the mean f-score and the bars on each side illustrate the standard deviation. It should be noted that in the \emph{g:emb/d:ce} and \emph{g:emb/d:emb} cases the std bars are not visible due to tiny variations among different runs.}
  \label{fig:loss_stability}
\end{figure}

Table~\ref{table:stability_and_loss} compares the use of embedding/cross entropy as different choices for adversarial loss term for training of the generator and the discriminator networks.
To compare the stability of the training, statistics over validation accuracies are reported.
Fig.~\ref{fig:loss_stability} furthermore illustrates the validation set F-score mean, and standard deviation over 5 training runs.
These results show that using the embedding loss for the generator makes GAN training stable.
We observed similar behavior when training with other hyper-parameters.

\setlength{\tabcolsep}{4pt}
\begin{table}
  \begin{center}
    \caption{TuSimple validation set accuracy statistics over different training iterations (every 10K), comparing the stability of different choices for adversarial losses}
    \label{table:stability_and_loss}
    \begin{tabular}{cc|ccc|c}
      \hline\noalign{\smallskip}
      \multicolumn{2}{c}{Loss} & \multicolumn{3}{c}{Accuracy statistics:} & \\
      Generator & Discriminator & mean & var & max & Equations \\
      \noalign{\smallskip}\hline\noalign{\smallskip}
      Cross entropy & Cross entropy & 33.84 & 511.71 & 58.11 & \ref{eq:lgen_regular} and \ref{eq:ldisc_ce} \\
      Cross entropy & Embedding & 0.00 & \textbf{0.00} & 0.02 & \ref{eq:lgen_regular} and \ref{eq:ldisc_emb} \\
      Embedding & Cross entropy & 93.97 & 0.459 & 94.65 & \ref{eq:lgen_twoinputs}, \ref{eq:lgen_embeddingloss} and \ref{eq:ldisc_ce} \\
      Embedding & Embedding & \textbf{94.17} & 0.429 & \textbf{94.98} & \ref{eq:lgen_twoinputs}, \ref{eq:lgen_embeddingloss} and \ref{eq:ldisc_emb} \\
      \noalign{\smallskip}\hline
    \end{tabular}
  \end{center}
\end{table}
\setlength{\tabcolsep}{1.4pt}

\begin{figure}[t]
  \centering
  \includegraphics[width=0.95\columnwidth]{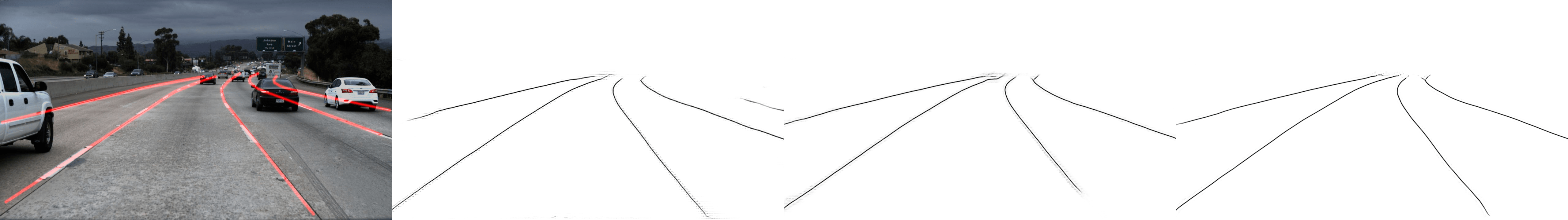}
  \includegraphics[width=0.95\columnwidth]{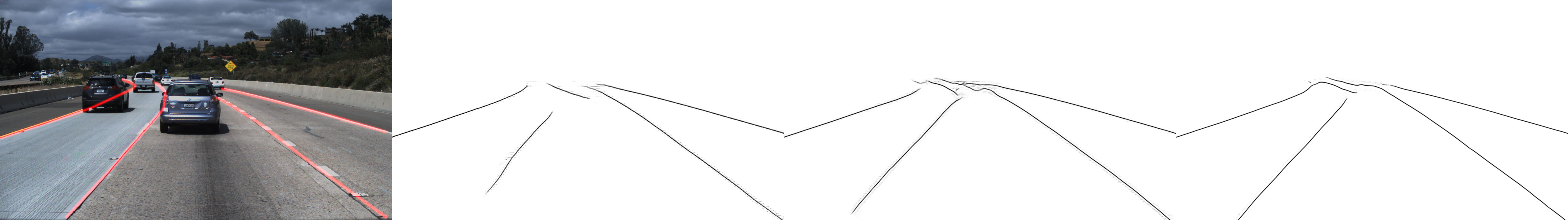}
  \includegraphics[width=0.95\columnwidth]{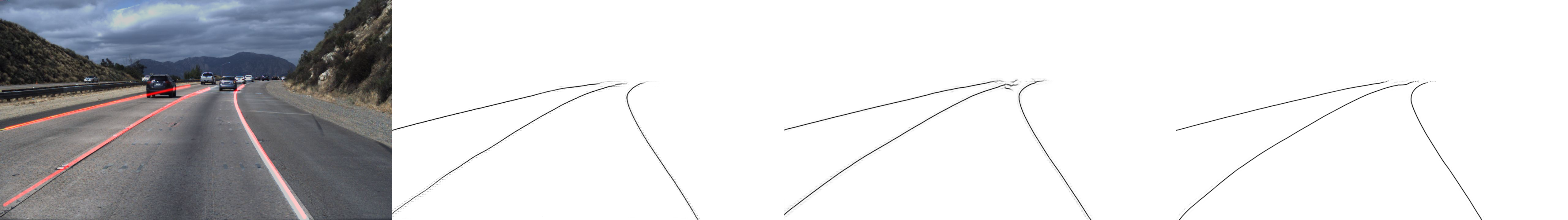}\\
  {\hspace{0.5em} label on data \hspace{3em} EL after DB 3 \hspace{3em} EL after DB 5 \hspace{3em} EL after DB 7 \hspace{0.5em}}
  \caption{Comparison of taking the embedding loss (EL) after a particular dense block (DB): the data and the label (left) and the prediction results of the different settings (right three images).
  Details are best viewed on a computer screen when zoomed in
  }
  \label{fig:embedding_layer}
\end{figure}

The features used for the embedding loss can be taken at different locations in the discriminator.
In this section we explore three options: taking the features either after the 3rd, 5th, or 7th dense block.
We note that the 3rd block contains the first shared convolution layers with both the image input and the predictions or labels, and that the 7th block contains the final set of convolutions before the classifier of the network.
Results for the TuSimple lane marking detection validation set are given in Table~\ref{table:embedding_layer} and in Fig.~\ref{fig:embedding_layer}.
From the results, we conclude that the later we take the embeddings, the better the score and the more similar the predictions are to the labels.

\setlength{\tabcolsep}{4pt}
\begin{table}
  \begin{center}
    \caption{Ablation study on embedding extraction layer}
    \label{table:embedding_layer}
    \begin{tabular}{l|ccc}
      \hline\noalign{\smallskip}
      Embedding loss after block \# & Accuracy (\%) & FP & FN \\
      \noalign{\smallskip}\hline\noalign{\smallskip}
      Dense block 3 (first block after joining) & 93.91 & 0.1013 & 0.1060 \\
      Dense block 5 & 94.01 & 0.0733 & 0.0878 \\
      Dense block 7 (before classifier) & \textbf{94.94} & \textbf{0.0592} & \textbf{0.0673} \\
      \noalign{\smallskip}\hline
    \end{tabular}
  \end{center}
\end{table}
\setlength{\tabcolsep}{1.4pt}

\section{Discussion}
\subsection{Comparison with Other Lane Marking Detection Methods}
\label{sec:tusimple_comparison}

In Table~\ref{table:tusimple_simple_leaderboard} we showed the results on the TuSimple lane marking data set with \elgan{} ranking 4th on the leaderboard.
In this section, we compare our method in more detail to the other two published methods: Pan et al.~\cite{pan2017spatial} (ranking 2nd) and Neven et al.~\cite{neven2018towards} (ranking 5th).

Neven et al.~\cite{neven2018towards} argue in their work that post-processing techniques such as curve fitting are preferably not done on the output of the network, but rather in a birds-eye perspective.
To this extent they train a separate network to learn a homography to find a perspective transform for which curve fitting is easier.
In our work we show that it is possible to achieve comparable accuracy results without having to perform curve fitting at all, thus omitting the requirement for training and evaluating a separate network for this purpose.

Pan et al.~\cite{pan2017spatial} use a multi-class approach to lane marking detection, in which each lane marking is a separate class.
Although this eases post-processing, it requires more complexity in label creation and makes the task more difficult for the network: it should now also learn which lane is which, requiring a larger field of view and yielding ambiguities at lane changes.
In contrast, with our GAN approach, we can learn a simpler single-class problem without requiring complex post-processing to separate individual markings.
Pan et al.~\cite{pan2017spatial} also argue that problems such as lane marking detection can benefit from spatial consistency and message passing before the final predictions are made.
For this reason they propose to feed the output of a regular segmentation network into a problem specific `spatial CNN' with message passing convolutions in different directions.
This does indeed result in a better accuracy on the TuSimple data set compared to \elgan{}, however, it is unclear how much is attributed to their spatial CNN and how much to the fact that they train on a non-public data set which is 20 times larger than the regular TuSimple data set.


\subsection{Analysis of the Ablation Study}

As we observed in the comparison of the different adversarial loss terms as presented in Table~\ref{table:stability_and_loss} and Fig.~\ref{fig:loss_stability}, using the embedding loss for the generator makes the training more stable and prevents collapses. 
The embedding loss, in contrast to the usual formulation with the cross entropy loss, provides stronger signals as it leverages the existing ground-truth rather than basing it only on the discriminator's internal representations of fake-ness and plausibility.

Therefore, using a normal cross entropy loss can result in collapses, in which the generator starts to explore samples in the feature space where the discriminator's fake/real comprehension is not well formed. 
In contrast, using the embedding loss, such noise productions result in high differences in the embedding space and is strictly penalized by the embedding loss.
Furthermore, having an overwhelming discriminator that can perfectly distinguish the fake and real distributions results in training collapses and instability.
Hence, using an embedding loss with better gradients that flow back to the generator likely results in a more competent generator.
Similarly, it is no surprise that using an embedding loss for the discriminator and not for the generator results in a badly diverging behavior due to a much more dominating discriminator and a generator that is not penalized much for producing noise.

In the second ablation study, as presented in Table~\ref{table:embedding_layer} and Fig.~\ref{fig:embedding_layer}, we observed that using deeper representations for extracting the embeddings results in better performance.
This is perhaps due to a larger receptive field of the embeddings that better enables the generator to improve on the higher-level qualities and consistencies.

\subsection{GANs for Semantic Segmentation}

\begin{figure}[!t]
  \centering
  \includegraphics[width=0.18\columnwidth]{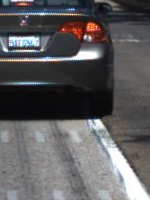}
  \includegraphics[width=0.18\columnwidth]{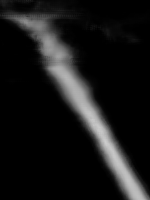}
  \includegraphics[width=0.18\columnwidth]{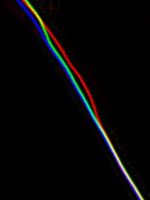}
  \\
  {data close-up \hspace{1em} regular CNN \hspace{1em} 3$\times$ \elgan{}}
  \caption{Example detail of input data (left), a regular semantic segmentation output (center), and three different \elgan{} models trained with the same settings shown as red, green, and blue channels (right)}
  \label{fig:discussion_illustration}
\end{figure}

Looking more closely at the comparison between a regular CNN and \elgan{} (Fig.~\ref{fig:tusimple_simple_pp}), we see a distinct difference in the nature of their output.
The non-GAN network produces a probabilistic output with a probability per class per pixel, while \elgan{}'s output is similar to a possible label, without expressing any uncertainty.
One might argue that the lack of being able to express uncertainty hinders further post-processing.
However, the first step of commonly applied post-processing schemes is removing the probabilities by thresholding or applying argmax (e.g.~\cite{neven2018towards,pan2017spatial}).
In addition, the independent per-pixel probabilistic output of the regular CNN might hide inter-pixel correlation necessary for correct post-processing. 
The cross entropy loss pushes the network to output a segmentation distribution that does not lie on the manifold of possible labels.

In \elgan{} and other GANs for semantic segmentation, networks are trained to output a sample of the distribution of possible labels conditioned on the input image.
An example is shown in Fig.~\ref{fig:discussion_illustration}, from which we clearly see the selection of a sample once the lane marking is occluded and the network becomes more uncertain.
Although this sacrifices the possibility to express uncertainty, we argue that the fact that it lies on, or close to, the manifold of possible labels, it can make post-processing easier and more accurate.
For the task of lane marking detection we indeed have shown that the semantic segmentation does not need to output probabilities.
However, for other applications this might not be the case.
A straightforward approach to re-introduce expressing uncertainty by a GAN, would be to simply run it multiple times conditioned on extra random input or use an ensemble of \elgan{}s.
The resulting samples which model the probability on the manifold of possible labels would then be the input to post-processing.


\section{Conclusions}

In this paper, we proposed, studied and compared \elgan{} as a method to preserve label-resembling qualities in the predictions of the network. 
We showed that using \elgan{} results in a more stable adversarial training process. 
Furthermore, we achieved state-of-the-art results on the TuSimple challenge, without using any extra data or complicated hand-engineered post-processing pipelines, as opposed to the other competitive methods.


\section*{Acknowledgments}

The authors would like to thank Nicolau Leal Werneck, Stefano Secondo, Jihong Ju, Yu Wang, Sindi Shkodrani and Bram Beernink for their contributions and valuable feedback.


\bibliographystyle{splncs}
\bibliography{bibliography}


\newpage
\appendix
\section*{Appendix A: \\ Network Architecture and Training Configuration}

Following are details on the hyper-parameters to ensure reproducibility.
We first list the network architecture set-up used for our experiments in more detail:

\begin{itemize}
  \item \emph{Generator}: Architecture: Tiramisu DenseNet~\cite{jegou2017the}, number of dense blocks in down/up sampling paths: 7, number of 3$\times$3 conv layers in each dense block: [1, 2, 3, 4, 6, 8, 8], growth-rate: 18, non-linearity: ReLU, initialization: He~\cite{he2015delving} , dropout rate: 0.1.
  \item \emph{Discriminator}: Architecture: two-headed DenseNet~\cite{huang2016densely}, joining the two heads: concatenation after the second dense block, number of dense blocks: 7, number of 3$\times$3 conv layers in each dense block: [1, 2, 3, 4, 6, 8, 8], growth-rate: 8, non-linearity: ELU~\cite{clevert2015fast}, no dropout, embeddings taken from layer: after 7th dense block.
\end{itemize}

\noindent
The detailed training hyper-parameters are as follows:

\begin{itemize}
    \item \emph{General}: number of iterations: 150K, batch size: 8, training schedule: (300: disc, 200: gen).
    \item \emph{Generator}: optimizer: (Adam~\cite{kingma2014adam}, momentum: 0.9), learning rate: (exponential, init: 5e-4, decay power: 0.99, decay rate: 200), $L_2$ regularization scale: 1e-4, pre-training: 100K iterations.
    \item \emph{Discriminator}: optimizer: vanilla SGD, learning rate: (exponential, init: 1e-5, decay power: 0.99, decay rate: 800), pre-training: 10K iterations, $L_2$ regularization scale: 1e-5, adversarial loss $\lambda$ : 1.
\end{itemize}


\end{document}